\documentclass{article}

\usepackage{microtype}
\usepackage{graphicx}
\usepackage{xspace}
\usepackage{subfigure}
\usepackage{booktabs} 

\usepackage{hyperref}

\usepackage[accepted]{icml2023}

\usepackage{amsmath}
\usepackage{amssymb}
\usepackage{mathtools}
\usepackage{amsthm}

\usepackage[capitalize,noabbrev]{cleveref}

\theoremstyle{plain}

\theoremstyle{definition}

\theoremstyle{remark}

\usepackage[textsize=tiny]{todonotes}

\icmltitlerunning{Green Federated Learning}

\begin{document}

\twocolumn[
\icmltitle{Green Federated Learning}



\icmlsetsymbol{equal}{*}

\begin{icmlauthorlist}
\icmlauthor{Ashkan Yousefpour}{meta}
\icmlauthor{Shen Guo}{meta}
\icmlauthor{Ashish Shenoy}{meta}
\icmlauthor{Sayan Ghosh}{meta}
\icmlauthor{Pierre Stock}{meta}
\icmlauthor{Kiwan Maeng}{meta,psu}
\icmlauthor{Schalk-Willem Krüger}{meta}
\icmlauthor{Michael Rabbat}{meta}
\icmlauthor{Carole-Jean Wu}{meta}
\icmlauthor{Ilya Mironov}{meta}
\end{icmlauthorlist}

\icmlaffiliation{meta}{Meta}
\icmlaffiliation{psu}{Pennsylvania State University. Work done when Kiwan Maeng was a postdoc researcher at Meta.}

\icmlcorrespondingauthor{Ashkan Yousefpour}{yousefpour@meta.com}

\icmlkeywords{Federated Learning, Green AI}

\vskip 0.3in
]



\printAffiliationsAndNotice{} 

\newcommand{\carbon}{CO$_2$e\xspace} 
\newcommand{\papaya}{{\sc Papaya}\xspace}

\begin{abstract}
The rapid progress of AI is fueled by increasingly large and computationally intensive machine learning models and datasets. As a consequence, the amount of computing used in training state-of-the-art models is increasing exponentially (doubling every 10 months between 2015 and 2022), resulting in a large carbon footprint. Federated Learning (FL) --- a collaborative machine learning technique for training a centralized model using data of decentralized entities --- can also be resource-intensive and have a significant carbon footprint, particularly when deployed at scale. Unlike centralized AI that can reliably tap into renewables at strategically placed data centers, cross-device FL may leverage as many as hundreds of millions of globally distributed end-user devices with diverse energy sources. Green AI is a novel and important research area where carbon footprint is regarded as an evaluation criterion for AI, alongside accuracy, convergence speed, and other metrics. 


In this paper, we propose the concept of Green FL, which involves optimizing FL parameters and making design choices to minimize carbon emissions consistent with competitive performance and training time. The contributions of this work are two-fold. First, we adopt a data-driven approach to quantify the carbon emissions of FL by directly measuring real-world at-scale FL tasks running on millions of phones. Second, we present challenges, guidelines, and lessons from studying the trade-off between energy efficiency, performance, and time-to-train in a production FL system. Our findings offer valuable insights into how FL can reduce its carbon footprint, providing a foundation for future research in the area of Green AI.

\end{abstract}

\section{Introduction}
\label{sec:introduction}


\emph{Federated learning} (FL) is a distributed learning paradigm where a large number of client devices, such as smartphones, collectively train a machine learning model using data located on client devices.
User data remains on client devices, and only updates to the model are aggregated within a centralized model at the server.
FL has emerged as a practical privacy-enhancing technology for on-device learning~\cite{fl_open_problems}.
Many real-world models have been trained using FL, including language models for predictive keyboards on Google Pixel, Apple's iOS, and Meta's Quest~\cite{google_fl, apple_fl_eval_p13n,oculus_fl}, Siri personalization~\cite{apple-fl-siri}, advertising, messaging, and search on LinkedIn~\cite{linkedin-fl}. 
 

While FL --- when coupled with technologies such as secure aggregation and differential privacy \cite{fl_open_problems, bonawitz2016practical, fedbuff, mcmahan2017learning} --- can be a practical solution to enhance user privacy, the training process in FL can result in non-negligible carbon emissions.
A recent study has shown that training a model with FL can produce as much as 80 kilograms of carbon dioxide equivalent (\carbon), exceeding that of training a higher capacity model, a large transformer, in the centralized training setting using AI accelerators~\cite{wu2022sustainable}. The relative inefficiency is attributable to several factors, including the overhead of training using a large collection of highly heterogeneous client hardware, additional cost for communication, and often slower convergence. 

Federated Learning's global carbon footprint is expected to increase as the industry increasingly adopts FL and more machine learning tasks shift away from the centralized setting. 
This is especially concerning since renewable sources of electricity may not be available in all locations, making Green FL a challenging goal to achieve~\cite{wu2022sustainable, fl_carbon}. \textit{Taking advantage of opportunities for efficiency optimization in FL is of paramount importance to make on-device learning greener.}

Recently, there has been growing interest in quantifying and reducing the carbon emissions of machine learning (ML) training and inference in the datacenter setting~\cite{ml_carbon_dodge, ml_carbon_strubell, ml_carbon_patterson,  lacoste2019quantifying, naidu2021towards}. However, the carbon footprint of Federated Learning (FL) and the factors that contribute to carbon efficiency in FL have yet to be thoroughly explored. Prior works have offer preliminary findings, either quantified the carbon effects of FL only in a simulation setting or with several simplifying assumptions~\cite{fl_carbon, wu2022sustainable}, offering only a partial picture.
These works focused on measurements and opportunity sizing, and restrained from exploring dimensions of the design space toward realizing Green FL.

\begin{figure}[t]
\centering
\includegraphics[width=0.95\columnwidth]{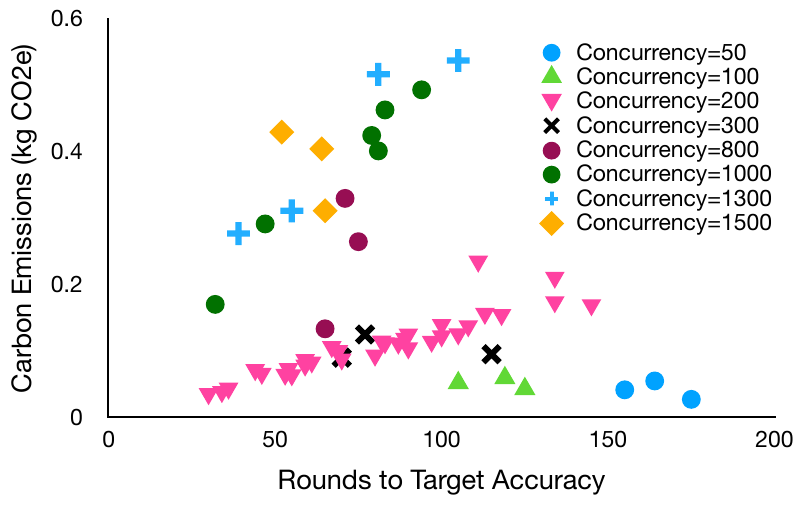}
\vspace{-0.25cm}
\caption{Carbon emissions of (synchronous) FL: the more rounds is required to reach a target accuracy and the higher the number of users active in training (i.e., \textit{concurrency}), the higher is the carbon emissions. Each point represents a training run with a different hyper-parameter (grouped by concurrency with marker colors and symbols). The graph shows the carbon emissions (Y axis) and the rounds to reach a target accuracy (X axis) for a language modeling FL task.}
\label{fig:CO2-FL}
\end{figure}

This paper presents a holistic carbon footprint analysis of a real-world FL task running on \papaya \cite{papaya}, a production FL system that operates on hundreds of millions of clients. This is the first study that provides a comprehensive view of Green FL by characterizing the emission profile of all major components involved, including the emissions from clients, the server, and the communication channels in between. To this end, we instrument and profile all major components of the FL system.

An important finding of our analysis is that \textbf{the carbon footprint of an FL task is highly correlated (among all parameters and artifacts) with the product of its running time and the number of users active in training (i.e., \textit{concurrency})}. We discuss this in more detail in Section~\ref{sec:impact}. We also provide an in-depth analysis of the multi-criterion optimization between carbon emissions, time to convergence, and training error.

Figure~\ref{fig:CO2-FL} presents results of measuring a production FL task for a range of hyperparameters. We can see that the number of rounds and concurrency are both positively correlated with the carbon footprint, and keeping one of these parameters constant, the relationship is nearly linear (see, for instance, the line corresponding to concurrency set to 200). These points become more evident throughout the paper. Other findings are the following:
\begin{itemize}
\item Compute on client devices, and the communication between the clients and the server are responsible for the majority of FL's overall carbon emissions (97\%). The carbon footprint attributable to the server-side computation is small ($\sim$1--2\%), while client computation is almost half of the contribution ($\sim$46--50\%). Upload and download networking costs are approximately 27--29\% and 22--24\%, respectively.
\item Asynchronous FL (FedBuff~\cite{fedbuff}) is faster than synchronous FL (FedAvg) as it advances the model more frequently in the face of stragglers, but it comes at the cost of higher carbon emissions.
\item Different training configurations that achieve similar model accuracy can have substantially different carbon impact, by up to 200$\times$, demonstrating the importance of hyper-parameter optimization.
\item To minimize the carbon footprint of FL, reduce training time, and achieve a high model quality, FL developers must focus on lowering the training time, e.g., through the right choice of the optimizer, learning rates, and batch sizes, while keeping the concurrency small.
\item Carbon footprint of a single language modeling FL task running for a few days at scale is of the order of 5--20 kg \carbon, similar to that of producing 1 kilogram of chicken \cite{food}. In practice, we expect model exploration, hyperparameter tuning, and incremental re-training to increase this multiple-fold (in the order of several tons of \carbon).
\end{itemize}

\subsection{Contributions}
We believe that distilling the complicated design space of the Green FL into several intuitive rules of thumb constitutes a valuable contribution. The rules and findings, even though may appear straight-forward in retrospect, are actionable and validated on a production platform. To the best of our knowledge, this is the first study to measure carbon emissions at scale for an industrial FL system across a range of hyperparameters. Before this study, there had been only intuitions and hypotheses. Our findings can help identify challenges and encourage further research towards the development of more sustainable and environmentally friendly FL systems.
Our main contributions can be summarized as follows:
\begin{itemize}
    \item We present a comprehensive evaluation of the carbon emission of a full production FL system stack by presenting the emission profile of all major components involved, including the emissions from clients, the server, and the communication channels in between. No prior work has done a carbon measurement study on a real-world FL production system at scale.
    \item Our empirical observations lead us to propose a set of key findings for Green FL, which identify the levers that have the most significant influence on the carbon footprint of~FL. 
    \item We propose a model that predicts the carbon footprint of an FL task prior to actual deployment.
    \item We show that using our recipe for Green FL, we can reduce the carbon footprint of FL training pipelines by as much as 200$\times$ while achieving similar model quality performance.
\end{itemize}

\paragraph{Disclaimer:} This study presents an estimation of carbon emissions based on other assumptions; measuring the actual energy and carbon emissions is indeed harder. Moreover, we only consider a production language modeling FL task. Given the state of FL deployment, we foresee language models to be responsible for a disproportionately large share of the total carbon footprint. While we believe the overall trend of the results should hold for other modalities or models, small variation in the findings and results may be expected.

\section{Why Green Federated Learning?}

Climate change is a pressing global issue believed to be caused by human activities such as burning fossil fuels, deforestation, and agriculture, which all emit greenhouse gasses (e.g., CO$_2$ and methane). Climate change has significant impacts on human communities, as well as on ecosystems and biodiversity. Mitigating harmful emissions is essential in addressing climate change~\cite{un_climate_change, masson2021climate}.

Green AI is the use of AI techniques and technologies in a way to reduce their environmental impact and promote sustainability in AI \cite{greenai, tackling}. Some examples are developing more energy-efficient algorithms and hardware, and reducing the carbon footprint of data centers. With the rapid growth of AI (e.g., the amount of compute for training state-of-the-art models doubled every 10 months between 2015 and 2022~\cite{three_eras}), it is imperative to understand the environmental implications, challenges, and opportunities of AI. By making AI more sustainable, we can reduce its environmental impact while also reaping the benefits that AI has to offer.

A related line of work addresses communication efficiency or model compression for FL ~\cite{konen2016federated,vogels2019powersgd,jiang2019model,Fetchsgd}. Historically, the primary objectives of these techniques have been cost reduction and not carbon emission savings per se -- closely related but different. Quantifying and reducing the carbon emissions of FL is our primary objective.

Although one might argue that renewable energy can power centralized AI systems \cite{google_cloud_sust, meta-pue, amazon_sust}, providing FL with renewable energy is inherently more challenging, as end-user devices are tied to their local energy mixes whose carbon footprint must be taken into account. In this paper we set out to study the problem of Green FL, present challenges, guidelines, and the lessons learned from realizing the trade-off between energy efficiency, performance, and time to train in a production FL system.

\section{Industry-Scale Federated Learning}

\subsection{Federated Learning Platform}
 Our company's production FL stack is built based on \papaya~\cite{papaya}, a recently proposed system for running federated learning and analytics tasks across millions of user devices. \papaya comprises two major subsystems: a server application that runs on a data center server and a client application that runs on end-user devices. In this study, we set out to measure the energy consumption and carbon footprint of both client-side and server-side resources used during training of a model in the federated learning system \papaya.
 
\begin{figure}[t]
\centering
\includegraphics[width=0.65\columnwidth]{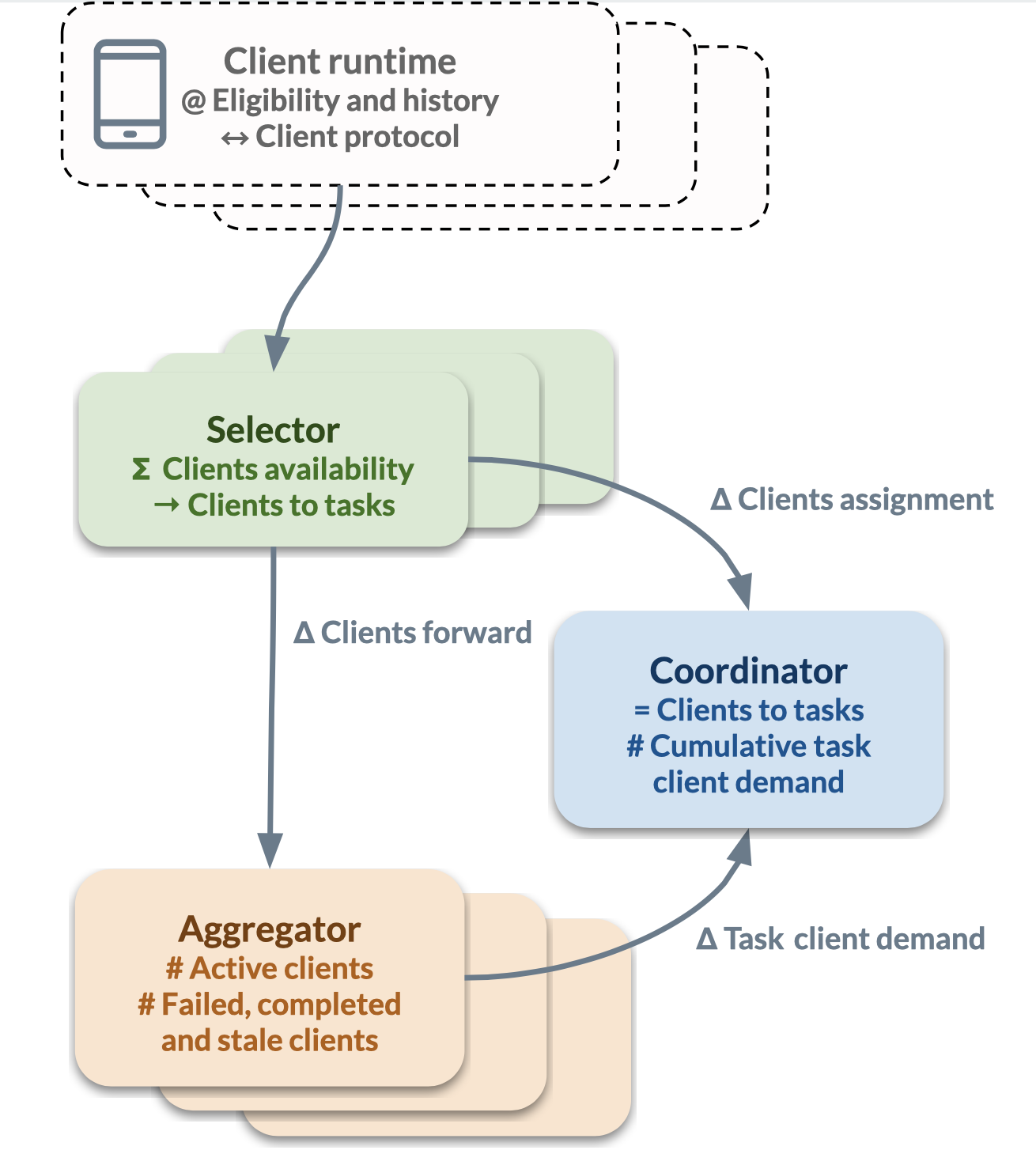}
\caption{Overall architecture of our production FL stack, based on \papaya \cite{papaya}.}
\label{fig:FL-platform}
\end{figure}

 The overall architecture of \papaya is presented in Figure~\ref{fig:FL-platform}. The  \papaya server has three main components: Coordinator, Selector, and Aggregator. There is one Coordinator, and the number of Selectors and Aggregators scales elastically based on the workload demand. The Coordinator assigns FL tasks to Aggregators based on load and assigns clients to tasks based on demand. 
Selectors report available clients and route clients to their assigned aggregator. 
Aggregators execute the client protocol, aggregate updates, and optimize the FL model. Regarding energy and carbon footprint, Aggregators and Selectors are responsible for the majority of processing and heavy lifting. The Coordinator is responsible for assigning FL tasks to Aggregators and clients to FL tasks, and centralized coordination.

\paragraph{Concurrency vs.~aggregation goal.} A device must meet a defined set of criteria to participate in FL training. Eligible devices report their availability to the Coordinator, which subsequently selects a subset of available devices for training. \emph{Concurrency} is the maximum number of clients that can train simultaneously. The \emph{aggregation goal} denotes the minimum number of client responses that must be received at the server before it updates the model.

In this study, for synchronous FL, the baseline implementation is FedAvg, whereas for asynchronous FL, it is FedBuff \cite{fedbuff}. In either settings, Adam is used as the server optimizer. In \emph{asynchronous} FL, based on the FedBuff protocol, a new device is immediately selected for training as soon as the server receives a client response. Therefore, the number of devices training at any given time essentially equals to the concurrency. Once the aggregation goal is met, the server model is updated, and clients selected thereafter receive the updated model. However, clients chosen earlier may still be training using the previous version of the server model, leading to a phenomenon called \emph{staleness}~\cite{fedbuff}. A 4-min timeout is imposed to limit the client training time and removing stragglers \cite{papaya}.

In contrast, \emph{synchronous} FL~\cite{mcmahan2017learning} proceeds in discrete rounds. At the beginning of a round, the server distributes the same model to a number of devices equal to the concurrency. At the end of the round, the server updates its model if it has received updates from at least as many users as the aggregation goal; it is worth noting that users may drop out during the round due to various reasons (such as the device no longer being idle or connected to Wi-Fi). In synchronous FL, the concurrency is also referred to as \emph{users per round}, and it is usually greater than the aggregation goal (a process called ``over-selection'') to account for the possibility of devices dropping out mid-round~\cite{bonawitz2019towards}.



\subsection{Large-scale FL Task: Language Modeling}

In all experiments for this study, we train a character-aware language model for a next word prediction task, similar to Kim et al.~\cite{char_aware_lm}. This model computes the probability of a sequence of words $S = w_1,\dots,w_T$ autoregressively as:
\begin{equation*}
p(S) =\prod_{i=1}^T{p(w_i|w_{<i})}.
\end{equation*}
More specifically, we use a character-level CNN with multiple filters, followed by a pooling layer that computes the final word embeddings. These are then encoded using a standard LSTM-based neural network that captures the sequential information in the input sequence. Finally, we use an MLP decoder followed by a \texttt{softmax} layer that converts the word-level outputs into final word-level probabilities over a fixed vocabulary. Using the notation where
\begin{itemize}
\item $i$ denotes the length of the sequence seen so far,
\item $x_{i,1}, x_{i,2},\dots, x_{i,L_i}$ are the characters of the $i$-th word in the input sequence,
\item $L_i$ is the length of the $i$-th word,
\item $e_{i}$ is the embedding for the $i$-th word,
\item $h_{i}$ is the hidden state,
\item $c_{i}$ is the state of the LSTM for the $i$-th word,
\item $W$ is the weight matrix,
\item $p(w_{i+1}|w_{\leq i})$ is the probability of the next word in the sequence computed using the MLP decoder and softmax layer,
\end{itemize}
then the model can be expressed as follows:
\begin{equation*}
\begin{aligned}
&e_i = \text{CNN}(x_{i,1}, x_{i,2},\dots, x_{i,L_i}) \\
&c_i, h_i = \text{LSTM}(h_{i-1}, c_{i-1}, e_i) \\
&p(w_{i+1}|w_{\leq i}) = \text{Softmax}(W^Th_i) \\
&\text{Perplexity}(w_0,w_1,\dots,w_i) = \left(\prod_{j=0}^{i-1} p(w_{j+1}|w_{\leq j})\right)^{-1/i}.
\end{aligned}
\end{equation*}
Perplexity measures the degree of uncertainty of a language model when it generates a new token averaged over sequence lengths. (Tokens are the basic units of text and can be characters, words, subwords, or other segments of text) Formally, perplexity is defined as the normalized inverse probability of sequences.

\paragraph{Number of devices.} The total pool of clients that can potentially participate in FL training can reach tens of millions. In each round, some devices (e.g., 1000) are selected to participate in the training (i.e., the {\em concurrency} parameter). One FL experiment can have hundreds or thousands of rounds, hence involving hundred thousands or millions of unique clients in a language modeling FL task running for a few days. 

Instead of using the users' data for mobile keyboard predictions, which could raise privacy concerns, we use publicly available, representative data downloaded to the user devices before training. We used \verb|pushift.io|’s Reddit FL benchmark dataset \cite{reddit_ds} in all experiments\footnote{Meta was not involved in the collection of data from Reddit.}. This dataset is publicly available, previously collected, and currently hosted by \verb|pushift.io|, consisting of user comments on \verb|reddit.com|. Thus this dataset has a natural non-IID partitioning and is representative of a real-world data distribution for mobile keyboard predictions. It also exhibits the archetypal power-law phenomenon of the number of comments per user. The dataset comprises millions of users, with an average of 34 samples per user. Each device participating in the FL is randomly assigned an anonymized user id from the \verb|pushift.io|’s Reddit dataset to use as their training data.

\paragraph{Stopping Criteria.}
We run an FL experiment until either the language model reaches a target perplexity on a hold-out test set, or a maximum time limit of 2 days is reached.
We set the target perplexity to be 175 or lower for our tasks and stop the task when the perplexity is at or lower than the target for five consecutive rounds. 

Due to the large number of experiments in this study, we set the target perplexity higher and the time limit shorter than those of the typical production models. The carbon emissions of the at-scale production models of the same task are expected to be roughly $10\times$ higher than the numbers reported in this study.

\subsection{Experiment Parameters}
We explored different settings of hyperparameters, separately optimizing for model performance, time to reach target accuracy, and carbon emission. We discuss these choices next.

For the optimizer running on the clients, we use SGD with no momentum. Alternatives (e.g., Adam) require additional on-device memory for the optimizer state (i.e., momentum buffers). Another important consideration is that in scenarios where clients possess limited data (which is often the case), they may not execute sufficient local steps to leverage the benefits of the momentum buffers. In such cases, the momentum-based optimization techniques may not be as effective. 

For the server optimizer, we wanted to be as general as possible, and we chose Adam for the server updates~\cite{adam}. Adam is more general than SGD or SGD with Momentum, and the parameters of Adam can be chosen to essentially replicate the performance of SGD or SGD with Momentum \cite{choi2019empirical}. On the server side in FL, we do not see any evidence that using a more compute-intensive optimizer has any significant impact on carbon emissions, although it should help the other dimensions --- reduce time to reach a target accuracy and improve overall accuracy. Our setup, the server updating the global model using the Adam optimizer and the clients using SGD, is called FedAdam~\cite{adam}.

 
We carefully evaluated hyperparameters for all applicable settings. We experimented with synchronous FL and asynchronous FL. For synchronous FL, the baseline implementation is FedAvg, whereas for asynchronous FL, it is FedBuff \cite{fedbuff}. However, since we use Adam as the server optimizer, both synchronous and asynchronous FL get the benefits of adaptive optimizers and perform better than their baselines. The hyperparameters are listed in Table~\ref{hyperparam_table}.

\begin{table}[t]
\small
\caption{Hyperparameters and their values explored in the experiments. Aggregation goal is expressed here as a percentage of concurrency.}
\begin{tabular}{@{}ll@{}}
\toprule
Hyperparameter       & Values                                         \\ \midrule
server learning rate & 0.0001, 0.001, 0.005, 0.01, 0.1, 1                    \\
client learning rate & 0.0001, 0.001, 0.01, 0.1, 0.5, 1                    \\
local epoch         & 1, 3, 5, 10, 15, 20                                \\
batch size           & 8, 16, 32                                      \\
Adam $\beta_1$           & 0.1, 0.5, 0.7, 0.9                             \\
Adam $\beta_2$           & 0.9, 0.99, 0.999                               \\
concurrency & 50, 100, 200, 300, 800, 1000, 1300, 1500             \\
aggregation goal     & {\scriptsize 8\%, 10\%, 25\%, 50\%, 65\%, 77\%, 80\%, 85\%, 100\%} \\
\bottomrule
\end{tabular}

\label{hyperparam_table}
\end{table}

\section{Measurement Methodology}

This section describes the measurement methodology we used to obtain our production FL stack's carbon emissions.

\subsection{End-user Device Resource Measurements}
\label{sec:client-measurement}
The FL software stack has a client runtime that executes on end-user devices for FL training tasks. To enable accurate measurement of compute time, upload, and download duration, we implemented a logger that records the vitals of the FL session, including the country from which the device is connected for the FL training, the model of the device, model download time, model upload time, and total duration of a single FL session. We use this information for power measurements of the devices. 

The logger records events happening on the production FL client runtime. The logger is based on a generic logging system used widely in our production client runtimes and has minimal resource footprint. The generic logging system guarantees that the events are defined, created, and processed consistently across all the apps and services. The logger runs in parallel with an FL session. The downstream of the logger is a server-side database to store the logs sent from the client runtime logger. 

\paragraph{Device training requirements.} 
It is a common practice in cross-device FL that training only takes place when the device is idle, charging, connected to an unmetered network (typically, Wi-Fi) \cite{fl_open_problems, google_fl, papaya}. 

\paragraph{Accounting for geography.} 
 Since the device is in the charging mode for FL, considering the source of energy for charging results in a more accurate carbon footprint estimate. To this end, we also consider the country where an end user connects from when the device is charging, since different countries have different carbon intensities. Carbon intensity reflects the amount of CO$_2$ emitted per unit of energy. We obtained country-level carbon intensities using the most recent reported year, e.g., 2020 or 2021, by Our World in Data~\cite{e2c_numbers}.

\paragraph{Optimization with regards to geography or heterogeneity.} 
One direction for reducing carbon impact of FL is to optimize carbon and performance with regards to the heterogeneity of clients (e.g., run more on clients with better power profile), or optimizing with respect to the carbon intensity of their location (e.g., run more on clients whose location has greener power source). These optimization strategies, although promising, may introduce bias and amplify unfairness due to the fact that clients with good energy profiles or with greener energy sources are underrepresented in the Global South. We encourage the community to study this direction further.

\paragraph{Power profile of phones.}
We acknowledge that power estimation of phones has been notoriously difficult \cite{oliner2013carat, couto2015greendroid}. Several methods exist to get the compute and communication power of end-user devices. One may try to approximate the power consumption of phones based on modeling different components, e.g., Wi-Fi units and TCP/IP layers~\cite{xiao2013modeling}, although these models may not be accurate. Another approach is to estimate the power drainage by looking at the phone's battery level over time. This method is a coarse-grained and noisy proxy to power consumption, as the battery life also depends on factors such as the age of the battery and ambient temperature. Moreover, the battery drain may differ for the same app usage across devices. Recent studies use collaborative methods for more accurate power measurements \cite{almeida2021smart, bustamante2022batterylab}. In this study, the main challenge for power estimation is the {\em diversity of devices} (see below). To address this challenge, we use Android phone's \emph{power profile}. The power profile is an XML file (typically named \verb|power_profile.xml|) that Android device manufacturers must provide to specify parameters of different electronic components and the approximate battery drainage caused by these components over time~\cite{android-power-profile}. This is the method we adopt as it provides accurate data for phone power consumption based on manufacturer information.

\paragraph{Diversity of Android devices.}
There are more than tens of thousands of distinct Android device models (also observed in \cite{wu2019machine, linkedin-fl}), and obtaining the power estimates for every device model that participated in our experiments would not be feasible. We instead focus on a subset of representative mobile phones---210 most commonly seen Android phones in the language modeling FL task in production. These devices represent more than 20\% of the total devices participating in the FL task in production. The power estimates for different components of these phones are measured by their manufacturers and are available from several sources~\cite{gh-pp-1, gh-pp-2, gh-pp-3, gh-pp-4}. 

We extract from \verb|power_profile.xml| the power consumption of the CPU and Wi-Fi components. In \papaya FL training of a language model is done on device CPU at present. (GPU support by PyTorch Mobile is largely limited to inference.) The listing below illustrates a snippet of a \texttt{power\_profile.xml} for Google Pixel 7.
{\scriptsize
\begin{verbatim}
<?xml version="1.0" encoding="utf-8"?>
<device name="Android">
    ... [text] ...
    <item name="screen.on">98</item>
    <item name="screen.full">470</item>
    <item name="modem.controller.sleep">2.5</item>
    <item name="modem.controller.idle">4.5</item>
    <item name="modem.controller.rx">169</item>
    ... [more text] ...
</device>
\end{verbatim}}


We impute values for phones with missing \verb|power_profile.xml| files using corresponding numbers from devices with the same SoC or similar phones with comparable characteristics. 

\paragraph{Wi-Fi Power.}
From the \verb|power_profile.xml| file, we use the fields \verb|wifi.active|, \verb|wifi.controller.rx|, \verb|wifi.controller.tx|, and \\
\verb|wifi.controller.voltage| to determine the communication power of the Wi-Fi, as these fields report the current and voltage when transmitting or receiving data \cite{android-power-values}. The receiving power of an end-user phone would be
\begin{equation*}
\label{eq:wifi-power}
    P_{\text{user\_rx}} = (I_{wa} + I_{wrx})\times V_{w}~,
\end{equation*}
where $I_{wa}$, $I_{wrx}$, $V_{w}$ denote \verb|wifi.active|, \verb|wifi.controller.rx|, and \verb|wifi.controller.voltage|, respectively. The transmission power is computed similarly with \verb|wifi.controller.tx| as $P_{\text{user\_tx}} = (I_{wa} + I_{wtx})\times V_{w}$.

\begin{figure}[t]
\centering
\includegraphics[width=1\columnwidth]{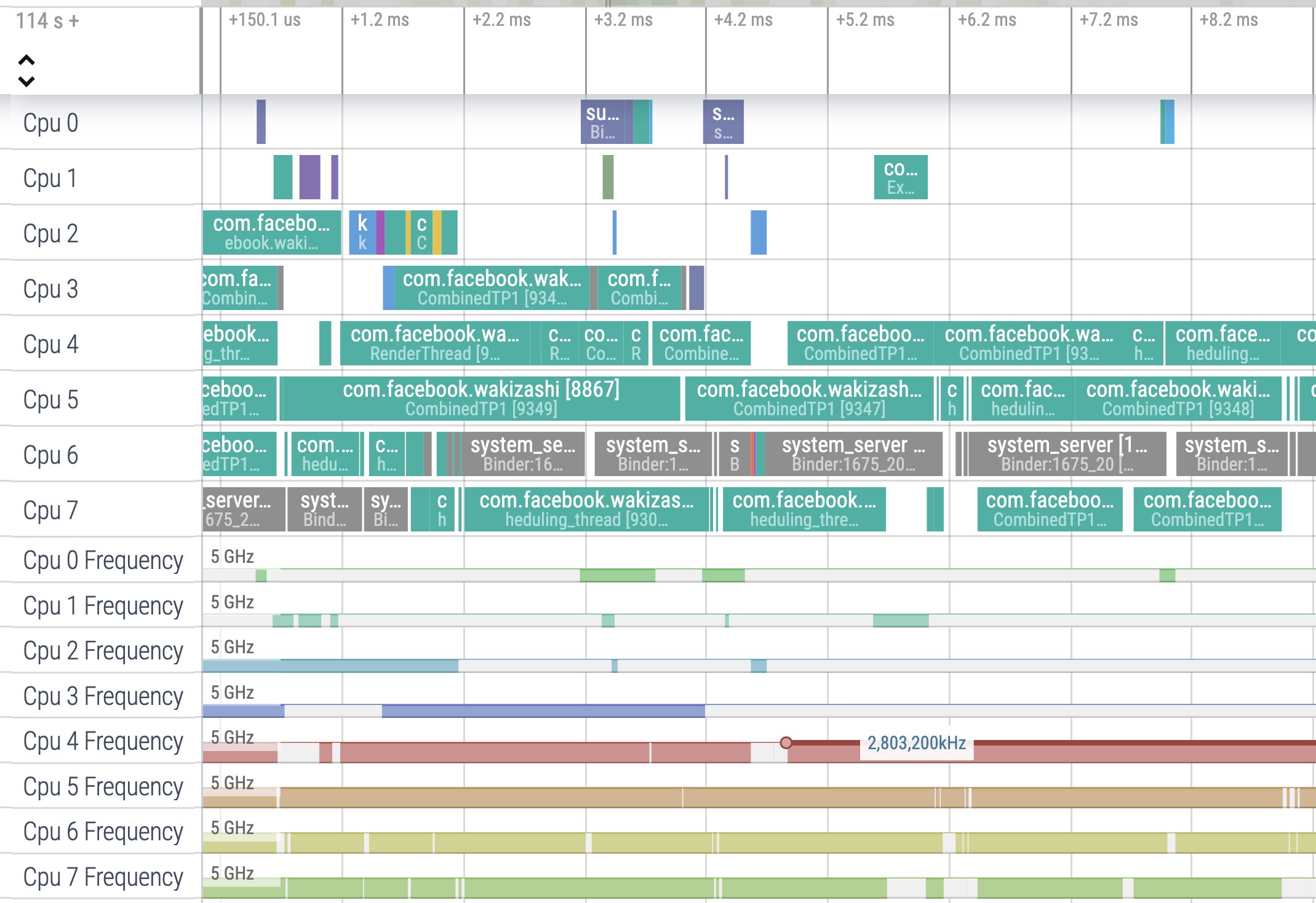}
\vspace{-0.25cm}
\caption{A snapshot of the 8 cores of the phone when running an FL task. Cores of the big cluster (CPUs 4 through 7) are running the FL task and are at the maximum frequency of 2.8GHz.}
\label{fig:cpu-full}
\end{figure}

\begin{figure}[t]
\centering
\includegraphics[width=1\columnwidth]{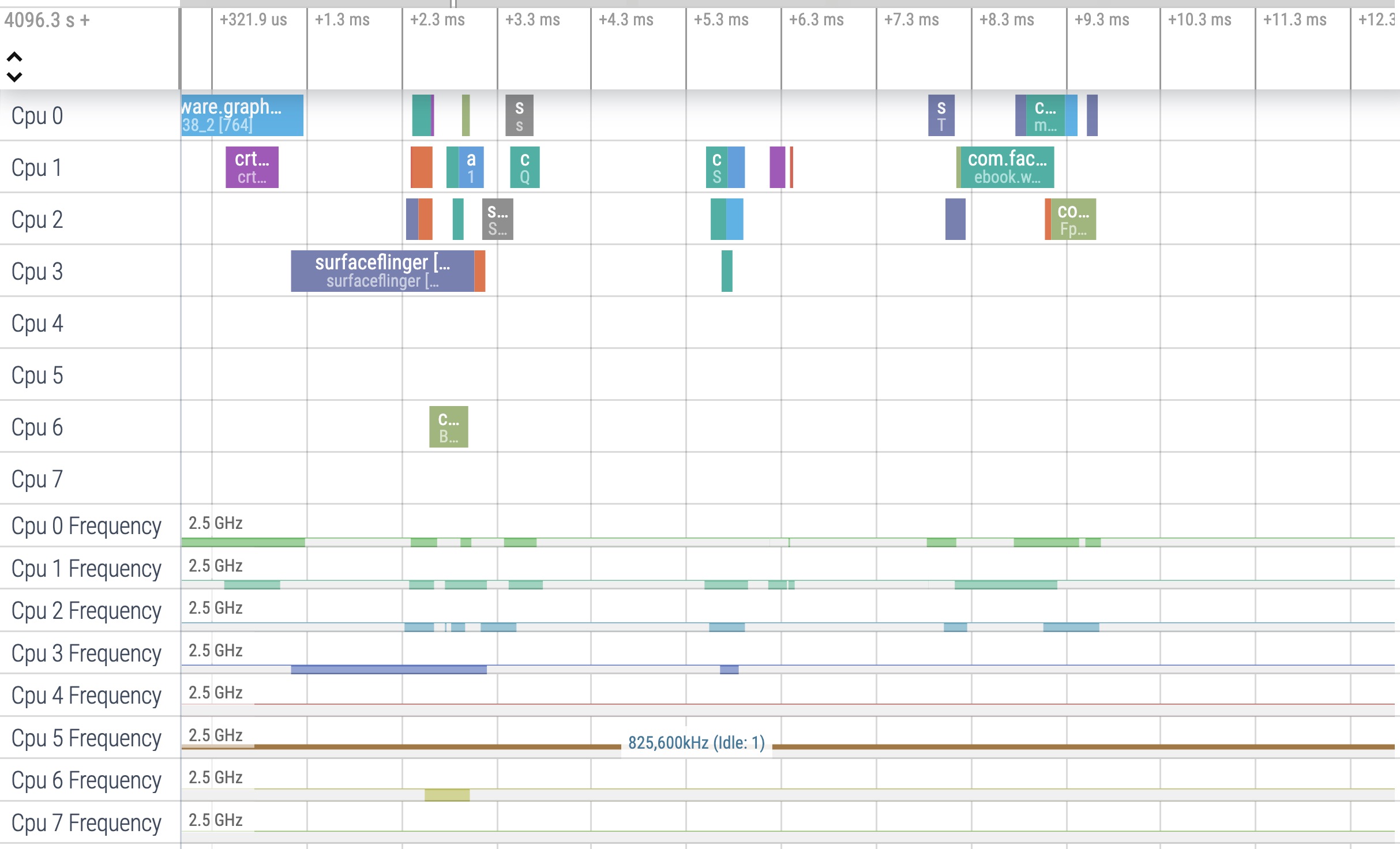}
\vspace{-0.25cm}
\caption{A snapshot of the 8 cores of the phone when it is idle. The cores of the big cluster are idle.}
\label{fig:cpu-idle}
\end{figure}

\paragraph{CPU Power.}
For estimating the CPU power of phones, we need to know the compute resource pattern of the language modeling task on the phones. We did a field study on a few phones running the FL language modeling task for this. We used the Perfetto tool for profiling and analyzing the resource usage trace and confirmed that the FL task runs when the device is idle, and it runs on the ``big'' cluster of the CPU. The following is a representative example. Google Pixel~3 based on the  Qualcomm SDM845 Snapdragon 845 SoC has two CPU clusters: a ``small'' cluster with four 1.8~GHz Kryo 385 cores for efficiency and a ``big'' cluster with four 2.8~GHz Kryo 385 cores for performance. Figure~\ref{fig:cpu-full} shows a snapshot of the 8 cores of the phone when running an FL task. Cores of the big cluster (cores 4 through 7) are running the FL task and are at the maximum frequency of 2.8GHz. Figure~\ref{fig:cpu-idle} confirms that when the phone is idle, the big cluster is idle and running at a lower frequency of 0.8GHz. 

The \verb|power_profile.xml| file has currents for all CPU clusters running at different frequencies. We find the total current by adding these values (concretely, \verb|cpu.cluster_power.cluster|,\\ \verb|cpu.active|, and \verb|cpu.core_power.cluster|) corresponding to the highest frequency belonging to the ``big'' cluster. Hence for FL training on the device, the current for CPU is the addition of the values in these 3 fields. We use Watt's law to convert current to power and assume that the phones operate at 3.8V~\cite{phone-voltage}. 

By multiplying the power and the FL session duration obtained from the logger -- namely upload time, download time, and processing time -- we can get the energy consumption for an FL session on a phone. We additionally confirm that the values resulting from this methodology are consistent with those reported in previous studies~\cite{wu2022sustainable, halpern2016mobile, kim2020autoscale}. Our methodology also accounts for the clients that drop out or time out during training, as their session information is reported in the logger.


\subsection{Server Resource Measurements}


We measure the carbon footprint of the server as follows. There are three main server components in the \papaya stack: Aggregator, Coordinator, and Selector. The most power-intensive computations happen in the Aggregator and Selector, while the Coordinator is responsible for matching FL tasks to clients and Aggregators and orchestration. To ensure accurate measurements of power consumption, we monitored the physical servers that run the FL task (Aggregator and Selector). We describe our methodology for measuring the carbon footprint of an individual task on these servers.

\paragraph{Aggregator.} To accurately estimate the carbon footprint of a single task on the physical servers, we measure the CPU utilization of the Aggregator during the execution of the language modeling FL task. We use the CPU utilization as a proxy for the power consumption specifically attributable to the FL task being executed on the server.

We consider the periods where the Aggregator runs only the language modeling FL task. First, we identify the Aggregator that runs a particular FL task. Next, we select a ``{\em stable}'' period when there is no failure, and the Aggregator is relatively underloaded. It is important to consider this period since the Coordinator reassigns FL tasks when it detects failed or overloaded Aggregators~\cite{papaya}, hence tracking an FL task would not be feasible. We observe that utilization of the Aggregator for the language modeling FL task is less than $1\%$, which also includes background processes. To get a conservative upper bound, we assume that server utilization is $1\%$ for the FL task. (Looking ahead, small errors in the estimate of the server utilization have a negligible impact on the results due to the small footprint of the server compute.)

Knowing the hardware specification of the physical servers running Aggregator, at $1\%$ utilization, we measured Aggregator's power consumption for running the language modeling FL task at $45$W. 
We multiply this number by the Power Usage Effectiveness (PUE) of our datacenters, 1.09, which accounts for the additional energy required to support the datacenter infrastructure (mainly cooling)~\cite{meta-pue}. 


Load balancing and other techniques of \papaya may impact where the Aggregator and Selector run. However, for this study, we assume they run uniformly across different datacenters.  
We use the weighted average carbon intensity model to account for the carbon intensity of different Meta datacenters that reside in different locations and regions \cite{meta-datacenters}. We obtain the weighted average of the carbon intensities of the countries where Meta datacenters are located, and the weight is the number of datacenters in that country. 

\paragraph{Selector.} Since most of the processing happens in the Aggregator, the Aggregator's carbon footprint dominates that of the Selector. We conservatively assume the same carbon footprint value for the Selector as for the Aggregator.

\subsection{Networking Infrastructure Resource Measurements}

For the networking and infrastructure resources, we adopt the standard methodology that considers all hardware assets on the path between the end-user and the FL server, namely, access, metro, edge, and core networks~\cite{vishwanath2015energy, baliga2010green, jalali2014energy, wu2022sustainable}. The access network is the first network user connects to, and it typically includes ADSL Ethernet, Wi-Fi access point, or 3G/4G/5G access point. The metro and edge network aggregate traffic from several users' access points, regulate access and usage, and represent the gateway to the global Internet, which consists of an edge Ethernet switch, broadband network gateways (BNGs), and edge routers \cite{vishwanath2015energy}. The core network, consisting of core routers, is the backbone of the Internet, connecting the metro and edge network to the datacenter. Schematically for cross-device FL we can have: client $\rightarrow$ Wi-Fi access point $\rightarrow$ edge Ethernet switch $\rightarrow$ BNG $\rightarrow$ edge routers $\rightarrow$ core routers $\rightarrow$ edge routers $\rightarrow$ data center Ethernet switch $\rightarrow$ data center.

In this setting, power consumption of the networking infrastructure connecting the end-user to the FL server in the datacenter can be obtained using the energy-per-bit model, as \cite{jalali2014energy, vishwanath2015energy}:
\begin{equation*}
    P_{\text{network}} = (E_a+E_{\text{as}}+E_{\text{bng}}+n_eE_e+n_cE_c+E_{\text{ds}})\times B,
\end{equation*} where $B$ is the bandwidth usage of the FL session, $n_e$ is the number of edge routers, $n_c$ is the number of core routers, and $E_a, E_{\text{as}}, E_{\text{bng}}$, $E_e, E_c, E_{\text{ds}}$ denote the energy per bit of the Wi-Fi access point, the edge Ethernet switch, the BNG, an edge router, a core router, and data center Ethernet switch respectively \cite{vishwanath2015energy}. We adopt constants from Vishwanath et al. \cite{vishwanath2015energy}. The bandwidth usage of the session, $B$, can be calculated using the model size divided by the upload or the download time.
\section{Carbon Emissions of FL}
\label{sec:impact}

In this section, we present the results of our study and measurements. We use CO$_2$-equivalents (\carbon), a standardized measure to express the global-warming potential of various greenhouse gases as a single metric. Carbon dioxide (CO$_2$) is not the sole greenhouse gas contributing to climate change. There exist other gases that also have a significant impact on the environment, and the aggregate effect of all these gases is quantified as \carbon: the number of metric tons of CO$_2$ emissions with the same global warming potential as one metric ton of another greenhouse~gas.

\begin{figure}[t]
\centering
\includegraphics[width=0.83\columnwidth]{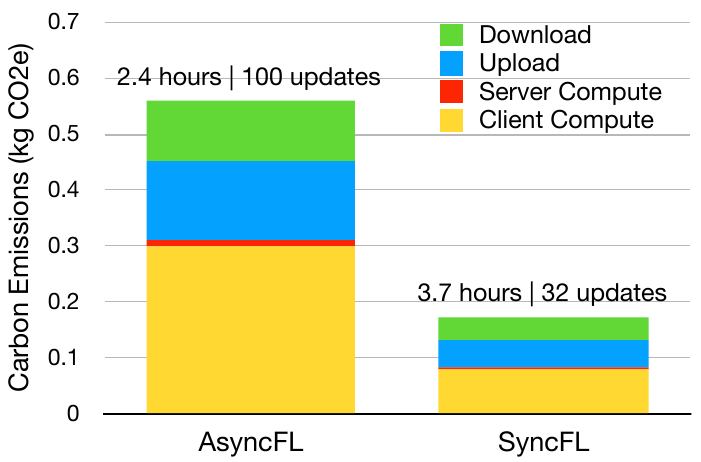}
\caption{Carbon emissions of SyncFL and AsyncFL to reach a target perplexity. The text above the bars shows the time and rounds it takes to reach the target perplexity.}
\label{fig:1}
\end{figure}

\subsection{Estimating Carbon Emissions of Industry-scale FL System}

We quantified the carbon emissions of our synchronous and asynchronous FL system through hundreds of experiments, each with a different set of hyperparameters. 

First, we illustrate the total carbon impact of synchronous and asynchronous FL. Figure \ref{fig:1} shows the carbon emissions of synchronous and asynchronous FL training in our production stack to reach a target perplexity of 175. We have tuned both methods by finding the choice of hyper-parameters that led to the lowest time to target perplexity, as also used in Huba et al.~\cite{papaya}. In this setup, concurrency and aggregation goal are both set to 1{,}000.

We can see that synchronous FL has a smaller carbon footprint compared to asynchronous FL. This is in contrast to the faster convergence of asynchronous FL (2.4 hours), which involves more model updates at the server (100). Asynchronous FL converges faster to the target perplexity due to its fast model updates. Due to its frequent iterations, asynchronous FL involves more clients. This result shows a fundamental trade-off between synchronous and asynchronous FL: if tuned well, \textbf{asynchronous FL is faster than synchronous FL as it advances the model more frequently in the presence of stragglers, but it comes at the cost of higher carbon emissions.}


We can also see that the majority of the carbon footprint is contributed by the client compute --- consistent with FL's pushing the AI processing to the edge of the network. The server compute is a small fraction of the carbon emissions as shown in Figure \ref{fig:1} and other experiments. We observe that client compute and the communication between the clients and the server are responsible for the greatest share of FL's overall carbon emissions (97\%). The carbon footprint from the server-side computation is small ($\sim$1--2\%), while client computation contributes to almost half of the overall carbon footprint ($\sim$46--50\%). Upload and download networking costs are approximately 27--29\% and 22--24\%, respectively. 

\begin{figure}[t]
\centering
\includegraphics[width=1\columnwidth]{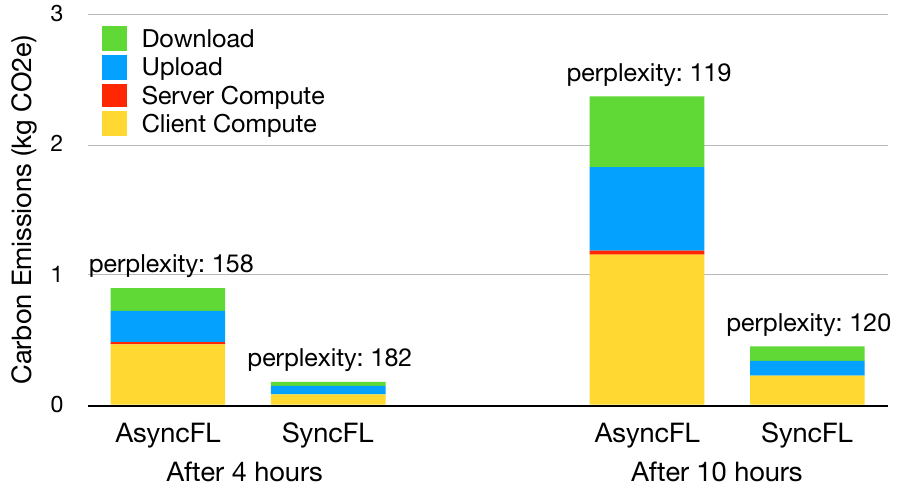}
\caption{Carbon emissions of SyncFL and AsyncFL after a fixed training time. The text above the bars shows the perplexity of the model at the specified time (lower is better).}
\label{fig:2}
\end{figure}

Figure \ref{fig:2} illustrates the carbon emissions of synchronous and asynchronous FL training in our production stack after a fixed time -- after 4 and 10 hours. In this experiment, instead of fixing the target perplexity and evaluating on training time, we fix the training time and measure the carbon emissions and the achieved perplexity (lower is better). The test perplexity is computed using data from 20 \emph{held-out} clients, to have quick evaluation. Each device has enough examples to have several hundreds of samples for evaluation. Because test perplexity with so few clients is noisy and can vary significantly from round to round, we smooth the test perplexity using an exponentially-weighted moving average with parameter $\alpha=0.3$ and declare that the test perplexity target has been reached when the smoothed test perplexity achieves the target. Asynchronous FL can advance the model faster and reach a lower perplexity at the cost of more carbon footprint. After 10 hours, synchronous FL is able to catch up to asynchronous FL with a similar perplexity of 120. The same contribution ratio among client compute, server compute, upload and download networking costs can be seen here~too.

In the rest of the experiments, we fix the target perplexity while evaluating carbon emissions and training time.

\subsection{Some Parameters Matter More}

\begin{figure}[b]
\centering
\includegraphics[width=0.85\columnwidth]{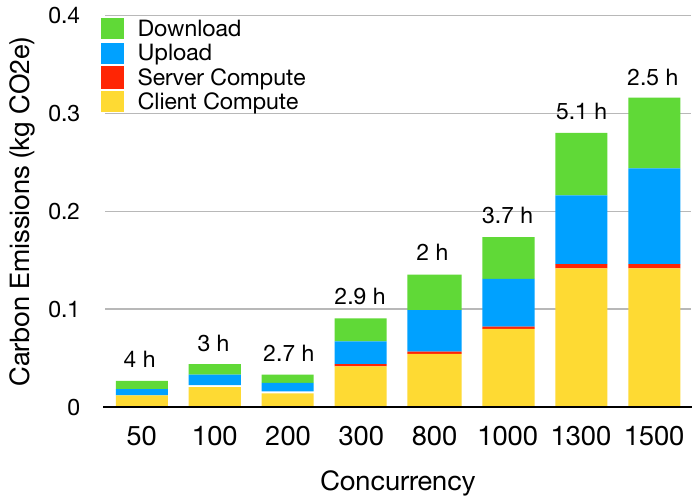}
\vspace{-0.25cm}
\caption{Higher concurrency leads to more carbon emissions. The numbers above bars are the time (in hours) it takes to reach the target accuracy.}
\label{fig:3}
\end{figure}

In our study on hyperparameters in FL tasks, we observed that some parameter choices have a greater impact on carbon footprint than others. Specifically, the parameter of \emph{concurrency} plays a significant role. The relationship between concurrency and carbon emissions in synchronous FL is depicted in Figure~\ref{fig:3}, where we observe that as concurrency increases, so does the carbon footprint. Higher concurrency leads to more devices training simultaneously, resulting in increased resource utilization and only partially offset with potentially faster convergence. We note that the time to reach a target accuracy decreases only up to concurrency of~800, illustrating diminishing returns in training speed. Diminishing returns in training speed as a function of increasing the number of clients training in parallel is analogous to a similar phenomenon in large-batch training~\cite{papaya, bonawitz2019towards, charles2021large}.

\begin{figure*}[t]
\centering
\includegraphics[width=0.95\textwidth]{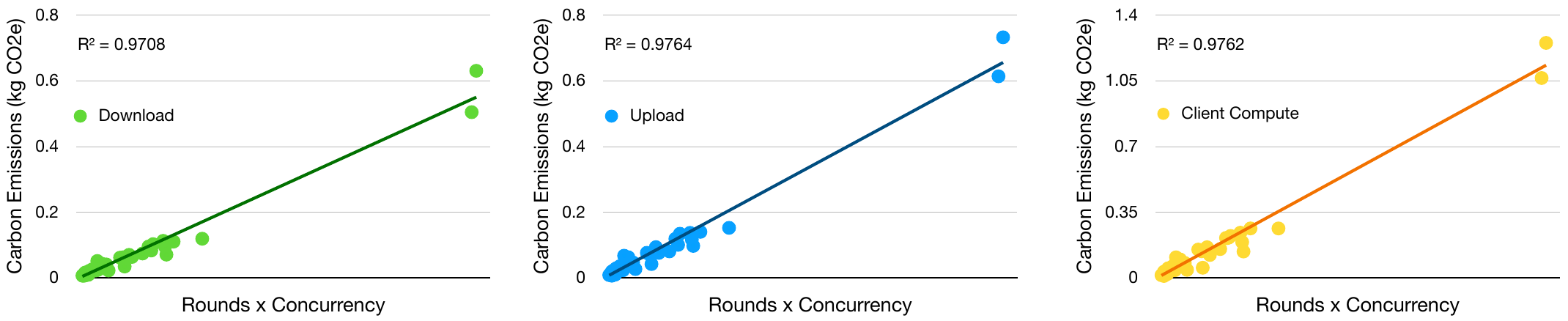}
\vspace{-0.25cm}
\caption{Carbon emission of synchronous FL is linearly correlated with the product of rounds it takes to reach a target accuracy and concurrency.}
\label{fig:8}
\end{figure*}

\begin{figure*}[t]
\centering
\includegraphics[width=0.95\textwidth]{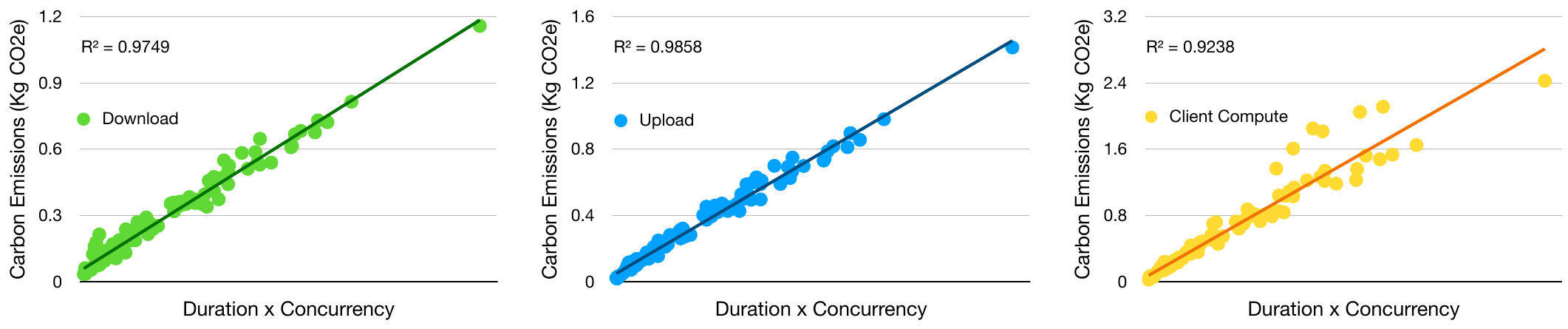}
\vspace{-0.25cm}
\caption{Carbon emission of asynchronous FL is linearly correlated with the product of the time it takes to reach a target accuracy and concurrency.}
\label{fig:9}
\end{figure*}

Among the parameters and artifacts of FL system design, we found that concurrency and time to reach a target accuracy, which translates to the number of rounds for synchronous FL and wall-clock time for asynchronous FL, have the most significant impact on carbon emissions. Conversely, other parameters such as learning rates, batch sizes, aggregation goals, and local epochs impact the convergence of the FL model towards the target accuracy, which in turn influences the time required to complete the training process. While these parameters do not directly affect carbon emissions, they do indirectly influence the overall training speed. Therefore, it is recommended that these parameters be included in the ``time'' and ``performance'' aspects of the multidimensional design in Green FL. On the other hand, concurrency impacts both time and carbon emissions directly and should be given greater consideration in FL system design for reducing carbon emissions.


Consistent with prior research \cite{papaya, bonawitz2019towards}, the present study indicates that larger values for local epoch do not yield improvements in training efficiency, particularly within the context of non-IID at-scale FL systems characterized by heterogeneous data and systems. On the contrary, larger local epoch values result in a marked increase in carbon emissions, largely attributable to the corresponding rise in client compute. Therefore, we recommend using smaller values for the local epoch, specifically in the 1 to 3 range.

\subsection{Predicting Carbon Emissions of FL}

We put forth a model of the relationship between time-to-convergence, model performance, and carbon emissions. By leveraging our model, FL practitioners can effectively forecast the carbon emissions of their system before initiating the training process.

We observed that concurrency is the most significant determinant of FL's carbon emissions. It has the largest effect on the resources, since concurrency most directly corresponds to the resource utilization of clients. While higher concurrency accelerates model convergence, it results in significantly higher carbon emissions. For instance, increasing concurrency by 10$\times$ increases the resource usage by 10$\times$ while only reducing the convergence time by 1.5$\times$ or 2$\times$. Therefore, the overall benefits of higher concurrency, considering resource consumption, do not scale linearly --- \textbf{increasing concurrency has diminishing returns considering convergence, model performance, and carbon emissions}. Higher concurrency reduces training duration, but increases resource usage even more.

To understand the relationship, we assume that the carbon emissions have a linear relationship with the product of concurrency and the number of rounds (or duration) it takes to reach a target accuracy. We validate our hypothesis in Figures~\ref{fig:8} and~\ref{fig:9}.

Figure~\ref{fig:8} shows the relationship between the product of rounds and concurrency and the carbon emissions for download, upload, and client compute in synchronous FL (carbon emissions of server compute is negligible). Different points on these scatter plots represent different training runs of the language modeling FL task. We use linear regression to find the fitting line that shows the aforementioned relationship. Figure~\ref{fig:8} also shows the $R^2$ values of the models, which is a goodness-of-fit measure for the linear regression models. We can see high $R^2$ values for the linear regression models, confirming \textbf{the product of rounds and concurrency is a good proxy to predict the carbon footprint of synchronous FL}.

Figure \ref{fig:9} shows a similar linear regression model for asynchronous FL. Since there is no concept of rounds in asynchronous FL, we treat duration (hours to reach a target accuracy) as an explanatory variable for the carbon footprint of asynchronous FL. We also see high goodness of fit ($R^2$ values) for these models. Hence, \textbf{the product of duration and concurrency is a good proxy to predict carbon footprint of asynchronous FL}. 

The carbon footprint of an FL is proportional to  concurrency and the number of rounds to convergence (or duration, in asynchronous FL). To estimate the carbon footprint prior to deployment, one needs to know their values as well as the coefficient of proportionality (i.e., the slope of the line in Figures~\ref{fig:8} and~\ref{fig:9}). Concurrency is a hyper-parameter of FL. For estimating rounds (or duration) to convergence, FL practitioners can use FL simulation tools, which is a common practice in the industry. The proportionality coefficient depends on multiple factors, such as the FL task, user population, and the FL infrastructure. In practice, one can estimate this coefficient from a few data points, i.e., by measuring carbon emissions of a task in several settings.

\begin{figure}[t]
\centering
\includegraphics[width=0.95\columnwidth]{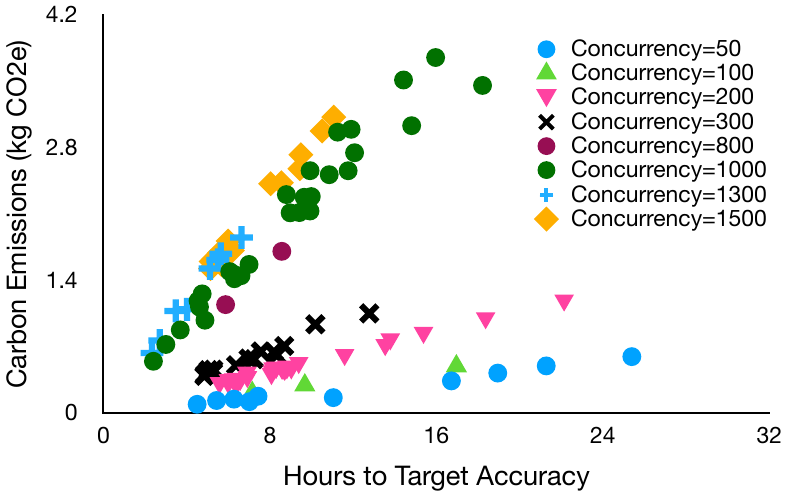}
\vspace{-0.25cm}
\caption{Carbon emissions of asynchronous FL increase linearly with the product of the time it takes to reach a target accuracy and concurrency. Each point represents a training run with a different hyper-parameter (grouped by concurrency with marker colors and symbols), its carbon emissions (Y axis, in kg \carbon) and the time it takes to reach a target accuracy (X axis, in hours). The more time is required to reach a target accuracy and the higher the concurrency, the higher is the carbon emission.}
\label{fig:11}
\end{figure}

In Figure~\ref{fig:11}, we present an overview of the design space for Green Federated Learning (FL) and highlight the trade-off between time, performance, and carbon emissions in asynchronous FL (the design space for synchronous FL was previously illustrated in Figure~\ref{fig:CO2-FL}). The scatter plot depicts various training runs of the FL task conducted through asynchronous FL, with different marker colors and symbols representing distinct concurrency values. Each point represents an experiment with a different hyper-parameter; we group the points by concurrency. We observe that the points corresponding to the same concurrency follow a linear trajectory, where higher concurrency leads to a steeper slope, implying a faster rate of \carbon accumulation. The cumulative carbon footprint of the task is a function of both its running time and the rate of carbon emission increase. Our analysis identifies concurrency and time to convergence as the two critical parameters for carbon emissions. While the former is under the direct control of the FL engineer, the latter is more indirect and reliant on the appropriate selection of hyperparameters. In particular, the high concurrency regime (which may be desirable, for instance, for its more robust privacy guarantees) puts a higher premium on hyperparameter tuning as longer training time translates into a larger carbon footprint. 

\section{Related Work}
To the best of our knowledge, our work is the first to conduct a large-scale carbon emission characterization of an industry-scale FL stack and different hyperparameters. We explore ways FL can be made more energy-efficient (``greener'') through the right selection of FL parameters and design choices. 

There has been growing interest in quantifying and reducing the carbon emission of machine learning (ML) training and inference in the datacenter~\cite{ml_carbon_dodge, ml_carbon_strubell, ml_carbon_patterson, anderson2022treehouse, naidu2021towards}. Nevertheless, the carbon footprint of FL has not yet been explored well. Prior works only quantify the carbon effect of FL in a simulation setting under several simplifying assumptions ~\cite{fl_carbon}. Other studies explore different ways for minimizing the energy footprint of client devices in Federated Learning~\cite{kim2021autofl, refl, kim2022fedgpo}, though not at large-scale scenarios like this study. Another work did a preliminary study of carbon emissions of FL~\cite{wu2022sustainable}; however, the authors also did not do their carbon emissions as comprehensively as our study does, as we log the FL session information and use the actual power measurements of the devices.

We take a data-driven approach to quantify carbon emissions of FL by directly measuring a real-world FL task at scale. We present challenges, guidelines, and lessons learned from studying the trade-off between energy efficiency, performance, and time to train in a production FL system.

Other related work could be the works on compression and quantization \cite{Fetchsgd, vogels2019powersgd}. Compressing the communications between the server and the clients could further reduce the carbon emissions of the FL training pipeline while presumably maintaining high model utility. For instance, we observed that the carbon emissions of communication in some settings could contribute to up to 60\% of the total emissions. Hence, reducing them by, say, a factor 4 with \texttt{int8} \cite{prasad2022reconciling} would reduce the total emissions by a factor of $1 / (.4 + 0.6 / 4) = 1.82$. 


\section{Conclusions and Future Work}

In this paper, we demonstrate how different FL parameters and design choices can impact the carbon footprint of a production FL system. Our empirical approach quantifies carbon emissions by directly profiling a real-world at-scale FL task. These measurements inform our guidelines and lessons learned on the trade-off between carbon emissions, target accuracy, and time to train in a production FL system.

We acknowledge that this study, like any, has some limitations. Recall that we used the power profiles of the 210 most commonly seen device models to obtain estimates of upload, download, and compute power for typical devices. We noted that these 210 devices represent 20\% of the total devices participating. The carbon emissions values we report based on these values are estimates. Although it is possible that the absolute carbon emissions would change if power profiles for additional devices were available, we believe that the same trends and overall conclusions hold.

As future research directions, we suggest investigating how compression and quantization techniques could apply to Green FL, potentially reducing the carbon footprint of the communication stack (at the expense of increasing client-side computations). Moreover, accounting for GPU or NPU resources is an interesting point when considering other FL tasks (e.g., computer vision) at scale. Alternative FL architectures, such as secure aggregation via cryptographic computations, federated ensemble learning, or federated split learning, present intriguing challenges as well. Moreover, unbiased optimization strategies with regards to geography and heterogeneity, while preserving fairness (as discussed in Section \ref{sec:client-measurement}) can be explored. Additionally, we encourage the research community to consider the impacts of differential privacy on the landscape of Green FL. Differential privacy would introduce privacy as an additional criterion, alongside accuracy, carbon, and time. Finally, we urge FL practitioners to consider the carbon footprint of their systems in their decision-making process.

\section*{Acknowledgements}
We would like to thank Pavel Ustinov, Kunal Bhalla, and Vlad Grytsun for the discussions and their invaluable help in this work; Harish Srinivas, Sam Lurye, Anthony Shoumikhin, Kaikai Wang, and Dweep Gogia for their continued support and for sharing their expertise; Karthik Prasad, Graham Cormode, Ramya Raghavendra, Arnald Shaju Samthambi, and John Nguyen for their insightful ideas and comments. We also like to thank Grace Nichols, Ajay Chand, Jesse He, Riten Gupta, Bharati Sethiya, and Gilbert Sanchez for helping us in mobile phone measurements.

\bibliography{reference}
\bibliographystyle{icml2023}



\end{document}